\documentclass[11pt]{article}

\usepackage{times}
\usepackage{latexsym}
\usepackage[T1]{fontenc}
\usepackage[utf8]{inputenc}
\usepackage{microtype}
\usepackage{inconsolata}
\usepackage{graphicx}
\usepackage{amsmath,amssymb}
\usepackage{booktabs}
\usepackage{multirow}
\usepackage{url}

\title{YFPO: Yoked Feature Preference Optimization with Neuron-Guided Rewards}

\newif\ifanonymous
\anonymousfalse

\ifanonymous
  \usepackage[review]{acl}   
\else
  \usepackage{acl}           
\fi

\ifanonymous
\author{Anonymous ACL submission}
\else
\author{
  Yifan Le \\
  Independent Researcher \\
  \texttt{leyifan@zju.edu.cn}
}
\fi

\begin{document}
\maketitle

\begin{abstract}
Preference optimization has become a widely used post-training paradigm for improving the reasoning abilities of large language models. Existing methods typically learn from preferred and dispreferred responses as external behavioral supervision, while largely ignoring capability-related signals encoded in the model's internal representations. In this work, we study whether such internal signals can provide useful auxiliary supervision for mathematical reasoning. We introduce \textbf{YFPO} (\textbf{Yoked Feature Preference Optimization}), a neuron-guided preference optimization framework that couples response-level preference learning with neuron-level rewards. YFPO first uses AttnLRP to identify math-associated internal features, and then derives an auxiliary reward from the activation margin of these neurons between preferred and dispreferred responses. This reward is combined with the standard preference optimization objective, encouraging the model to align external preferences with internal math-related features. We conduct small-scale experiments on GSM8K with a compact language model. Results show that neuron-guided rewards can influence preference optimization dynamics and yield measurable improvements in several settings, suggesting that internal representations can serve as lightweight and interpretable signals for reasoning-oriented post-training.
\end{abstract}

\begin{figure*}[t]
    \centering
    \includegraphics[width=1\textwidth]{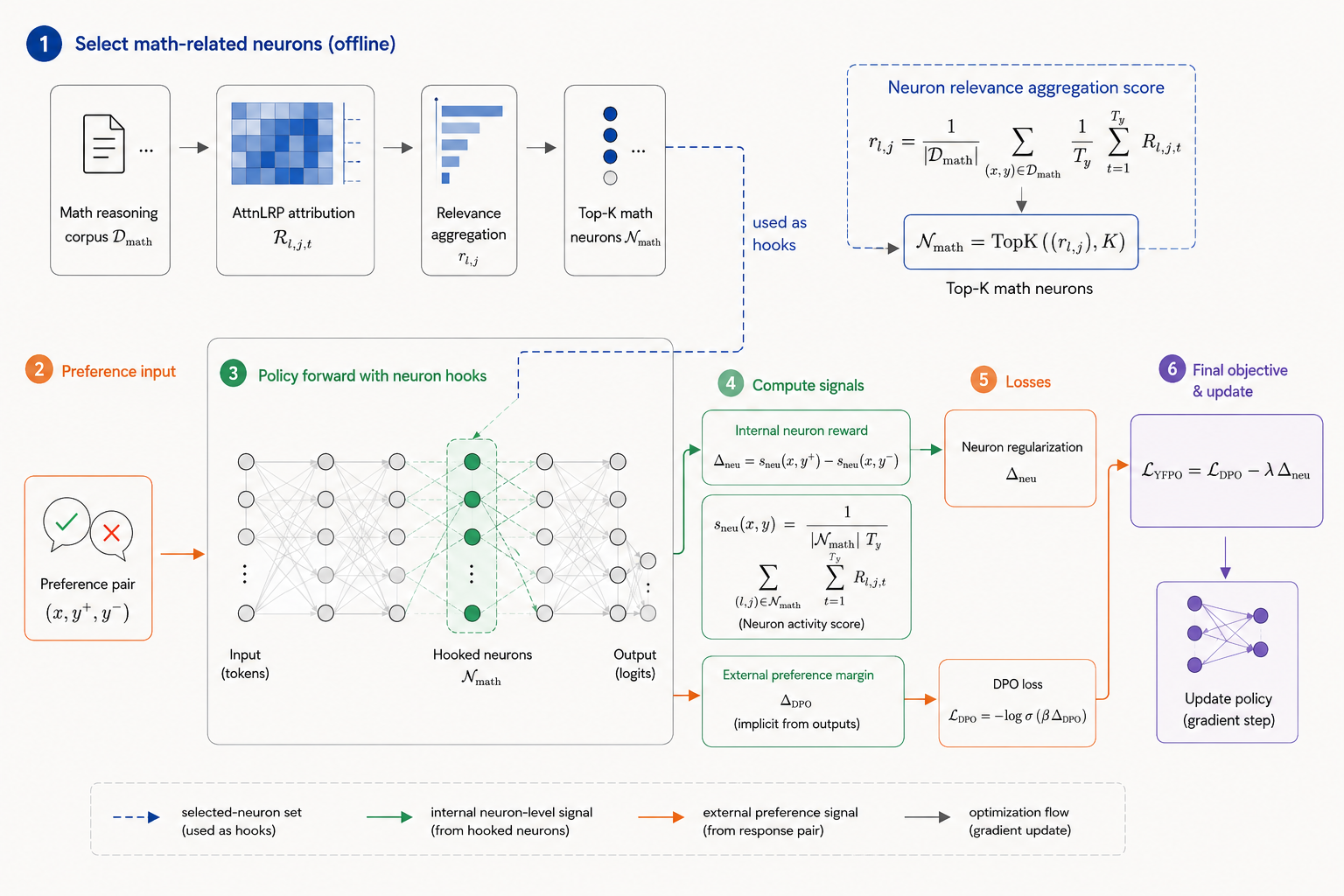}
    \caption{
    Overview of YFPO.
    In the offline stage, AttnLRP is used to identify a fixed set of top-$K$ math-related neurons.
    During preference optimization, YFPO extracts activations from these neurons for preferred and dispreferred responses, constructs a neuron reward margin, and combines it with the DPO objective.
    }
    \label{fig:yfpo_overview}
\end{figure*}

\section{Introduction}

Preference optimization has become an important post-training paradigm for improving large language models (LLMs), especially on reasoning-oriented tasks. Existing methods, including PPO-style RLHF and direct preference optimization methods such as DPO, IPO, KTO, ORPO, and SimPO, typically rely on external supervision signals such as human preferences, reward models, verifier feedback, or chosen/rejected response pairs \citep{schulman2017ppo,stiennon2020learning,ouyang2022training,rafailov2023dpo,azar2023ipo,ethayarajh2024kto,hong2024orpo,meng2024simpo}. For mathematical reasoning, such signals are often derived from final-answer correctness, process supervision, or automatic verifiers \citep{cobbe2021training,lightman2023lets,shao2024deepseekmath,DBLP:conf/nips/YuZZYZYDFLLLLLL25}. While effective, these objectives mainly tell the model which output is preferred, but rarely use internal computational signals involved in generating that output.

In this work, we ask whether internal model representations can provide useful auxiliary signals for preference optimization. Prior interpretability studies suggest that LLMs contain neuron-level structures associated with factual knowledge, language behavior, and task-specific patterns \citep{dai2022knowledge,tang2024language}. Attribution methods such as AttnLRP further enable relevance estimation over internal components \citep{DBLP:conf/icml/AchtibatHDJWLS24}. For mathematical reasoning, neurons with high relevance on math examples may provide coarse but informative signals about whether the model engages math-related internal features.

We introduce \textbf{YFPO} (\textbf{Yoked Feature Preference Optimization}), a neuron-guided preference optimization framework that augments DPO-style training with neuron-level rewards. YFPO first uses AttnLRP to identify a fixed set of math-related neurons. During preference optimization, it extracts their activations for preferred and dispreferred responses, constructs an auxiliary reward from the activation margin, and combines this reward with the standard DPO objective. As illustrated in Figure~\ref{fig:yfpo_overview}, this design uses the chosen/rejected structure of preference data to align external response-level preferences with internal math-related features.

Our contributions are threefold: (1) we study how internal math-related neuron signals can complement external preference data in preference optimization; (2) we propose YFPO, which converts AttnLRP-selected neuron activations into an auxiliary reward for DPO-style training; and (3) we conduct small-scale experiments on Qwen2-0.5B and GSM8K, showing that neuron-guided rewards can influence preference optimization dynamics and yield measurable improvements in several settings.

\section{Related Work}

\paragraph{Preference optimization and reasoning.}
Preference optimization has become a central technique for LLM post-training. Early RLHF methods typically use reward models and PPO-style reinforcement learning to align model outputs with human preferences \citep{schulman2017ppo,stiennon2020learning,ouyang2022training}. More recent direct preference optimization methods, including DPO, IPO, KTO, ORPO, and SimPO, simplify this pipeline by optimizing on chosen/rejected responses or preference-like labels without explicitly training a separate reward model \citep{rafailov2023dpo,azar2023ipo,ethayarajh2024kto,hong2024orpo,meng2024simpo}. For mathematical reasoning, preference optimization and reinforcement learning often rely on final-answer correctness, process supervision, verifiers, or preference data as external feedback signals \citep{cobbe2021training,lightman2023lets,shao2024deepseekmath,DBLP:conf/nips/YuZZYZYDFLLLLLL25}. In contrast, YFPO studies whether internal neuron-level signals associated with mathematical reasoning can provide auxiliary guidance during preference optimization.

\paragraph{Neuron-level interpretability.}
Neuron-level interpretability aims to identify internal components associated with specific knowledge, languages, or task behaviors. Prior work has used feature attribution, probing, activation statistics, and intervention to analyze neurons related to factual knowledge or language-specific behavior \citep{sundararajan2017axiomatic,belinkov2022probing,dai2022knowledge,tang2024language}. AttnLRP extends layer-wise relevance propagation to transformer models and enables efficient attribution over latent representations \citep{DBLP:conf/icml/AchtibatHDJWLS24}.  While these methods are commonly used to explain model behavior, YFPO uses relevance-selected math-related neurons as training signals: it converts their activation margin between preferred and dispreferred responses into an auxiliary reward, connecting internal capability representations with external preference learning.

\section{Method}

We propose \textbf{YFPO}, a neuron-guided preference optimization method that augments DPO-style training with an auxiliary reward from math-related neurons. YFPO contains two stages: offline neuron selection and online preference optimization. In the offline stage, we identify neurons associated with mathematical reasoning using AttnLRP. In the training stage, we compute a neuron reward margin between preferred and dispreferred responses and add it to the DPO objective.

\subsection{Math-Related Neuron Selection}

We use AttnLRP~\citep{DBLP:conf/icml/AchtibatHDJWLS24} to identify neurons associated with mathematical reasoning Given mathematical reasoning samples. AttnLRP assigns relevance scores to intermediate MLP channels according to their contribution to output tokens.

We use AttnLRP to select internal coordinates associated with
MLP down-projection modules, indexed by layer $l$ and coordinate $j$. For each neuron, we aggregate its signed relevance over mathematical samples and response tokens:

\begin{equation}
    r_{l,j} =
    \mathbb{E}_{x,y,t}
    \left[
    R_{l,j,t}(x,y)
    \right],
\end{equation}
where $R_{l,j,t}$ denotes the relevance assigned to neuron $(l,j)$ at response token $t$. We select the Top-$K$ neurons with the largest signed relevance scores as the AttnLRP-selected internal feature set:
\begin{equation}
    \mathcal{N}_{\text{math}}
    =
    \operatorname{TopK}_{(l,j)}
    \left(
    r_{l,j}
    \right).
\end{equation}
Unless otherwise stated, we use $K=1024$ and keep $\mathcal{N}_{\text{math}}$ fixed during preference optimization.

\subsection{Neuron-Guided Reward}

Given a preference example $(x,y^+,y^-)$, where $y^+$ and $y^-$ denote the preferred and dispreferred responses, we extract internal activations using forward hooks on the outputs
of the corresponding \texttt{mlp.down\_proj} modules. For a response $y$, we define its math-neuron score as:
\begin{equation}
    s_{\text{neu}}(x,y) =
    \mathbb{E}_{(l,j)\in \mathcal{N}_{\text{math}},\, t\in \mathcal{T}_{y}}
    \left[
    a_{l,j,t}(x,y)
    \right],
\end{equation}
where $a_{l,j,t}$ is the activation of neuron $(l,j)$ at response token $t$, and $\mathcal{T}_{y}$ denotes response tokens.
Before constructing the pairwise margin, sequence-level neuron scores
are standardized within the concatenated chosen/rejected batch:
\[
\tilde{s}_{\mathrm{neu}} =
\frac{s_{\mathrm{neu}}-\mu_{\mathcal{B}}}
{\sigma_{\mathcal{B}}+10^{-6}},
\]
where $\mu_{\mathcal{B}}$ and $\sigma_{\mathcal{B}}$ are the batch
mean and standard deviation.
The neuron reward margin is:
\begin{equation}
    \Delta_{\mathrm{neu}}
=
\tilde{s}_{\mathrm{neu}}(x,y^+)
-
\tilde{s}_{\mathrm{neu}}(x,y^-).
\end{equation}

A larger $\Delta_{\text{neu}}$ indicates that the preferred response obtains a higher activation score on the selected math-related neurons than the dispreferred response.

\subsection{YFPO Objective}

YFPO uses DPO~\citep{rafailov2023dpo} as the base objective. For a preference example $(x,y^+,y^-)$, the standard DPO margin is:
\begin{equation}
\begin{split}
    \Delta_{\text{DPO}} =
    &\left[\log \pi_{\theta}(y^+|x) - \log \pi_{\theta}(y^-|x)\right] \\
    &-
    \left[\log \pi_{\text{ref}}(y^+|x) - \log \pi_{\text{ref}}(y^-|x)\right].
\end{split}
\end{equation}
YFPO adds the neuron reward margin to the DPO loss:
\begin{equation}
    \mathcal{L}_{\text{YFPO}}
    =
    -\log \sigma(\beta \Delta_{\text{DPO}})
    -
    \lambda \Delta_{\text{neu}},
\end{equation}
where $\beta$ controls the DPO preference strength and $\lambda$ controls the neuron-guided reward strength. When $\lambda=0$, YFPO reduces to standard DPO. When $\lambda>0$, the model is optimized to prefer chosen responses while increasing their math-neuron activation margin over rejected responses.

Additional implementation details are provided in Appendix~\ref{app:implementation}.

\section{Experiments}

We conduct small-scale experiments to examine whether neuron-guided rewards can complement DPO-style preference optimization for mathematical reasoning.

\paragraph{Setup.}
We use Qwen2-0.5B as the base model and \texttt{xinlai/Math-Step-DPO-10K} as the preference dataset. We evaluate both the full 10K training set and a randomly sampled 2K subset. For YFPO, math-related neurons are selected offline using AttnLRP with $K=1024$, and the neuron reward margin $\Delta_{\mathrm{neu}}$ is added to the DPO objective. We compare standard DPO with DPO+YFPO under different $\lambda$ values. All models are evaluated on GSM8K using greedy decoding. The original Qwen2-0.5B obtains 0.3654 accuracy on GSM8K. Additional training-dynamics analysis and training details are provided in Appendices~\ref{app:training_dynamics} and~\ref{app:training_details}.

\begin{table}[t]
\centering
\small
\renewcommand{\arraystretch}{1.8}
\setlength{\tabcolsep}{6pt}
\begin{tabular}{lccc}
\toprule
Metric & DPO & YFPO$_{0.05}$ & YFPO$_{1.0}$ \\
\midrule
C1 & 0.3647 & 0.3738 & 0.3738 \\
C2 & 0.3662 & 0.3586 & 0.3669 \\
C3 & 0.3639 & 0.3707 & 0.3692 \\
\addlinespace[2pt]
Avg. & 0.3649 & 0.3677 & 0.3700 \\
Best & 0.3662 & 0.3738 & 0.3738 \\
\addlinespace[2pt]
$\Delta_{\mathrm{Avg.}}$ & -- & +0.0028 & +0.0051 \\
$\Delta_{\mathrm{Best}}$ & -- & +0.0076 & +0.0076 \\
\bottomrule
\end{tabular}
\caption{
GSM8K results on Qwen2-0.5B using the full 10K training set. YFPO$_{\lambda}$ denotes DPO+YFPO with coefficient $\lambda$. C1/C2/C3 denote three saved checkpoints.
}
\label{tab:gsm8k_10k}
\end{table}

\begin{table}[t]
\centering
\small
\renewcommand{\arraystretch}{1.8}
\setlength{\tabcolsep}{3.8pt}
\begin{tabular}{lccccc}
\toprule
Metric & DPO & YFPO$_{1.0}$ & YFPO$_{0.1}$ & YFPO$_{0.2}$ & YFPO$_{0.3}$ \\
\midrule
C1 & 0.3601 & 0.3616 & 0.3730 & 0.3616 & 0.3715 \\
C2 & 0.3647 & 0.3692 & 0.3609 & 0.3715 & 0.3669 \\
C3 & 0.3669 & 0.3624 & 0.3669 & 0.3624 & 0.3639 \\
\addlinespace[2pt]
Avg. & 0.3639 & 0.3644 & 0.3669 & 0.3652 & 0.3674 \\
Best & 0.3669 & 0.3692 & 0.3730 & 0.3715 & 0.3715 \\
\addlinespace[2pt]
$\Delta_{\mathrm{Avg.}}$ & -- & +0.0005 & +0.0030 & +0.0013 & +0.0035 \\
$\Delta_{\mathrm{Best}}$ & -- & +0.0023 & +0.0061 & +0.0046 & +0.0046 \\
\bottomrule
\end{tabular}
\caption{
GSM8K results on Qwen2-0.5B using a randomly sampled 2K subset. YFPO$_{\lambda}$ denotes DPO+YFPO with coefficient $\lambda$. C1/C2/C3 denote three saved checkpoints.
}
\label{tab:gsm8k_2k}
\end{table}

\paragraph{Results.}
Table~\ref{tab:gsm8k_10k} shows results on the full 10K training set. Standard DPO provides limited improvement over the original model, reaching a best accuracy of 0.3662. Adding YFPO improves the best checkpoint accuracy under both tested $\lambda$ values, with both settings reaching 0.3738. The $\lambda=1.0$ setting also improves average accuracy from 0.3649 to 0.3700.

Table~\ref{tab:gsm8k_2k} shows results on the 2K subset. YFPO improves the best checkpoint accuracy under all tested $\lambda$ values. The strongest best-checkpoint result is obtained with $\lambda=0.1$, reaching 0.3730 and improving over DPO by 0.0061. In terms of average accuracy, $\lambda=0.3$ performs best, improving from 0.3639 to 0.3674. These results suggest that math-related neuron rewards can provide useful auxiliary signals even under a lower-resource preference training setting.

\paragraph{Analysis.}
The results suggest that YFPO provides a useful auxiliary signal for preference optimization. On the full 10K set, YFPO improves the best checkpoint accuracy from 0.3662 to 0.3738, and the $\lambda=1.0$ setting improves the average accuracy from 0.3649 to 0.3700. On the 2K subset, YFPO improves best checkpoint accuracy under all tested $\lambda$ values, with the strongest result reaching 0.3730. These gains are modest but consistent across data-scale settings, indicating that math-related neuron activations can complement external chosen/rejected supervision.

However, YFPO does not follow a simple monotonic trend with respect to $\lambda$. Different reward strengths lead to different best checkpoints, showing that neuron-guided rewards are sensitive to reward scale and training stage. Appendix~\ref{app:training_dynamics} further shows that YFPO preserves the external DPO preference signal while introducing a distinct neuron-level signal. This suggests that the neuron reward is not merely a post-hoc explanatory metric, but an active training signal that can influence preference optimization dynamics.

\section{Conclusion}

We introduced YFPO, a neuron-guided preference optimization method that augments DPO-style training with rewards derived from math-related neurons. YFPO uses AttnLRP to identify relevant neurons offline and computes a neuron activation margin between preferred and dispreferred responses during training. In this way, YFPO connects external response-level preference supervision with internal capability-related signals.

Experiments on Qwen2-0.5B and GSM8K show that YFPO brings measurable improvements over standard DPO under multiple settings. On both the full 10K training set and the 2K subset, YFPO improves best-checkpoint accuracy, and in several cases also improves average checkpoint accuracy. These results indicate that activations of AttnLRP-selected internal features can provide useful auxiliary supervision.

Our analysis further suggests that neuron-guided rewards influence preference optimization dynamics while preserving the external DPO preference signal. Although the gains remain modest and sensitive to reward scale and checkpoint selection, YFPO provides measurable evidence that internal neuron-level signals can complement external preference data for reasoning-oriented post-training.
\section{Limitations}

This work has several limitations. First, our experiments are limited in scale. We mainly evaluate YFPO on Qwen2-0.5B and GSM8K, without systematic experiments on larger models, instruction-tuned models, or broader mathematical reasoning benchmarks. Therefore, the current results should be interpreted as initial evidence in a small-model setting rather than a general conclusion across model scales and tasks.

Second, the gains of YFPO are modest and checkpoint-sensitive. Although YFPO improves over DPO under several settings, its effectiveness depends on the reward coefficient $\lambda$, training stage, and data scale. Future work should conduct more systematic $\lambda$ sweeps, multi-seed experiments, and larger-scale evaluations to better assess the stability of neuron-guided rewards.

Third, YFPO relies on a fixed set of AttnLRP-selected neurons and uses their activation margin as an auxiliary reward. This design is simple and interpretable, but the selected neurons may change their role during training, and activation signals do not necessarily imply true causal contribution to reasoning ability. Future work should explore dynamic neuron selection, random-neuron baselines, causal intervention, and extensions to other post-training objectives such as SimPO, ORPO, GRPO, and process reward learning.
\bibliography{custom}

\begin{thebibliography}{17}
\providecommand{\natexlab}[1]{#1}

\bibitem[{Achtibat et~al.(2024)Achtibat, Hatefi, Dreyer, Jain, Wiegand, Lapuschkin, and Samek}]{DBLP:conf/icml/AchtibatHDJWLS24}
Reduan Achtibat, Sayed Mohammad~Vakilzadeh Hatefi, Maximilian Dreyer, Aakriti Jain, Thomas Wiegand, Sebastian Lapuschkin, and Wojciech Samek. 2024.
\newblock \href {https://proceedings.mlr.press/v235/achtibat24a.html} {Attnlrp: Attention-aware layer-wise relevance propagation for transformers}.
\newblock In \emph{Forty-first International Conference on Machine Learning, {ICML} 2024, Vienna, Austria, July 21-27, 2024}, volume 235 of \emph{Proceedings of Machine Learning Research}, pages 135--168. {PMLR} / OpenReview.net.

\bibitem[{Azar et~al.(2023)Azar, Rowland, Piot, Guo, Calandriello, Valko, and Munos}]{azar2023ipo}
Mohammad~Gheshlaghi Azar, Mark Rowland, Bilal Piot, Daniel Guo, Daniele Calandriello, Michal Valko, and R{\'e}mi Munos. 2023.
\newblock A general theoretical paradigm to understand learning from human preferences.
\newblock \emph{arXiv preprint arXiv:2310.12036}.

\bibitem[{Belinkov(2022)}]{belinkov2022probing}
Yonatan Belinkov. 2022.
\newblock Probing classifiers: Promises, shortcomings, and advances.
\newblock \emph{Computational Linguistics}, 48(1):207--219.

\bibitem[{Cobbe et~al.(2021)Cobbe, Kosaraju, Bavarian, Chen, Jun, Kaiser, Plappert, Tworek, Hilton, Nakano et~al.}]{cobbe2021training}
Karl Cobbe, Vineet Kosaraju, Mohammad Bavarian, Mark Chen, Heewoo Jun, Lukasz Kaiser, Matthias Plappert, Jerry Tworek, Jacob Hilton, Reiichiro Nakano, and 1 others. 2021.
\newblock Training verifiers to solve math word problems.
\newblock \emph{arXiv preprint arXiv:2110.14168}.

\bibitem[{Dai et~al.(2022)Dai, Dong, Hao, Sui, Chang, and Wei}]{dai2022knowledge}
Damai Dai, Li~Dong, Yaru Hao, Zhifang Sui, Baobao Chang, and Furu Wei. 2022.
\newblock Knowledge neurons in pretrained transformers.
\newblock In \emph{Proceedings of the 60th Annual Meeting of the Association for Computational Linguistics}.

\bibitem[{Ethayarajh et~al.(2024)Ethayarajh, Xu, Muennighoff, Jurafsky, and Kiela}]{ethayarajh2024kto}
Kawin Ethayarajh, Winnie Xu, Niklas Muennighoff, Dan Jurafsky, and Douwe Kiela. 2024.
\newblock Kto: Model alignment as prospect theoretic optimization.
\newblock \emph{International Conference on Machine Learning}.

\bibitem[{Hong et~al.(2024)Hong, Lee, and Thorne}]{hong2024orpo}
Jiwoo Hong, Noah Lee, and James Thorne. 2024.
\newblock Orpo: Monolithic preference optimization without reference model.
\newblock \emph{arXiv preprint arXiv:2403.07691}.

\bibitem[{Lightman et~al.(2023)Lightman, Kosaraju, Burda, Edwards, Baker, Lee, Leike, Schulman, Sutskever, and Cobbe}]{lightman2023lets}
Hunter Lightman, Vineet Kosaraju, Yura Burda, Harrison Edwards, Bowen Baker, Teddy Lee, Jan Leike, John Schulman, Ilya Sutskever, and Karl Cobbe. 2023.
\newblock Let's verify step by step.
\newblock \emph{arXiv preprint arXiv:2305.20050}.

\bibitem[{Meng et~al.(2024)Meng, Xia, and Chen}]{meng2024simpo}
Yu~Meng, Mengzhou Xia, and Danqi Chen. 2024.
\newblock Simpo: Simple preference optimization with a reference-free reward.
\newblock In \emph{Advances in Neural Information Processing Systems}.

\bibitem[{Ouyang et~al.(2022)Ouyang, Wu, Jiang, Almeida, Wainwright, Mishkin, Zhang, Agarwal, Slama, Ray et~al.}]{ouyang2022training}
Long Ouyang, Jeffrey Wu, Xu~Jiang, Diogo Almeida, Carroll~L. Wainwright, Pamela Mishkin, Chong Zhang, Sandhini Agarwal, Katarina Slama, Alex Ray, and 1 others. 2022.
\newblock Training language models to follow instructions with human feedback.
\newblock \emph{Advances in Neural Information Processing Systems}.

\bibitem[{Rafailov et~al.(2023)Rafailov, Sharma, Mitchell, Ermon, Manning, and Finn}]{rafailov2023dpo}
Rafael Rafailov, Archit Sharma, Eric Mitchell, Stefano Ermon, Christopher~D. Manning, and Chelsea Finn. 2023.
\newblock Direct preference optimization: Your language model is secretly a reward model.
\newblock \emph{Advances in Neural Information Processing Systems}.

\bibitem[{Schulman et~al.(2017)Schulman, Wolski, Dhariwal, Radford, and Klimov}]{schulman2017ppo}
John Schulman, Filip Wolski, Prafulla Dhariwal, Alec Radford, and Oleg Klimov. 2017.
\newblock Proximal policy optimization algorithms.
\newblock \emph{arXiv preprint arXiv:1707.06347}.

\bibitem[{Shao et~al.(2024)Shao, Wang, Zhu, Xu, Song, Bi, Zhang, Zhang, Li, Wu, and Guo}]{shao2024deepseekmath}
Zhihong Shao, Peiyi Wang, Qihao Zhu, Runxin Xu, Junxiao Song, Xiao Bi, Haowei Zhang, Mingchuan Zhang, Y.~K. Li, Y.~Wu, and Daya Guo. 2024.
\newblock Deepseekmath: Pushing the limits of mathematical reasoning in open language models.
\newblock \emph{arXiv preprint arXiv:2402.03300}.

\bibitem[{Stiennon et~al.(2020)Stiennon, Ouyang, Wu, Ziegler, Lowe, Voss, Radford, Amodei, and Christiano}]{stiennon2020learning}
Nisan Stiennon, Long Ouyang, Jeffrey Wu, Daniel~M. Ziegler, Ryan Lowe, Chelsea Voss, Alec Radford, Dario Amodei, and Paul~F. Christiano. 2020.
\newblock Learning to summarize with human feedback.
\newblock \emph{Advances in Neural Information Processing Systems}.

\bibitem[{Sundararajan et~al.(2017)Sundararajan, Taly, and Yan}]{sundararajan2017axiomatic}
Mukund Sundararajan, Ankur Taly, and Qiqi Yan. 2017.
\newblock Axiomatic attribution for deep networks.
\newblock \emph{International Conference on Machine Learning}.

\bibitem[{Tang et~al.(2024)Tang, Luo, Huang, Zhang, Wang, Zhao, Wei, and Wen}]{tang2024language}
Tianyi Tang, Wenyang Luo, Haoyang Huang, Dongdong Zhang, Xiaolei Wang, Xin Zhao, Furu Wei, and Ji-Rong Wen. 2024.
\newblock Language-specific neurons: The key to multilingual capabilities in large language models.
\newblock In \emph{Proceedings of the 62nd Annual Meeting of the Association for Computational Linguistics}.

\bibitem[{Yu et~al.(2025)Yu, Zhang, Zhu, Yuan, Zuo, Yue, Dai, Fan, Liu, Liu, Liu, Liu, Lin, Lin, Ma, Sheng, Tong, Zhang, Zhang, Zhang, Zhang, Zhu, Zhu, Chen, Chen, Wang, Yu, Song, Wei, Zhou, Liu, Ma, Zhang, Yan, Wu, and Wang}]{DBLP:conf/nips/YuZZYZYDFLLLLLL25}
Qiying Yu, Zheng Zhang, Ruofei Zhu, Yufeng Yuan, Xiaochen Zuo, Yu~Yue, Weinan Dai, Tiantian Fan, Gaohong Liu, Juncai Liu, Lingjun Liu, Xin Liu, Haibin Lin, Zhiqi Lin, Bole Ma, Guangming Sheng, Yuxuan Tong, Chi Zhang, Mofan Zhang, and 17 others. 2025.
\newblock \href {http://papers.nips.cc/paper\_files/paper/2025/hash/a4277440d50f1f15d2cb4c14f7e0c0d2-Abstract-Conference.html} {{DAPO:} an open-source {LLM} reinforcement learning system at scale}.
\newblock In \emph{Advances in Neural Information Processing Systems 38: Annual Conference on Neural Information Processing Systems 2025, NeurIPS 2025, San Diego, CA, USA, December 2-7, 2025 / Mexico City, Mexico, November 30 - December 5, 2025}.

\end{thebibliography}

\appendix

\section{Additional Implementation Details}
\label{app:implementation}

\paragraph{Hook placement.}
During YFPO training, forward hooks are applied to the output activations
of the \texttt{mlp.down\_proj} modules. For blocks represented in the fixed
AttnLRP index map, the indexed coordinates are used. For blocks without
indexed coordinates, the implementation uses the mean over the
\texttt{down\_proj} output coordinates as a deterministic fallback. This keeps the activation space used for neuron reward computation aligned with the neuron set selected by AttnLRP.
For blocks represented in the fixed AttnLRP index map, the indexed
coordinates are used. For blocks without indexed coordinates, the
implementation uses the mean over the \texttt{down\_proj} output
coordinates as a deterministic fallback.

\paragraph{Response-token masking.}
The neuron score is computed only over response tokens. Prompt tokens are excluded because the length and content of the input problem may otherwise dominate the activation score. We average over both selected neurons and response tokens to reduce sensitivity to sequence length and neuron count.

\paragraph{Signed relevance selection.}
YFPO uses the signed relevance values produced by AttnLRP for neuron selection. Specifically, neurons are selected according to their aggregated relevance scores $r_{l,j}$, rather than the absolute values $|r_{l,j}|$. Thus, the selected neurons are those with the largest positive relevance to mathematical reasoning outputs under the attribution procedure.

\paragraph{Activation-based reward.}
The neuron reward is computed from the raw activations of the selected neurons, without taking absolute values. This reward encourages preferred responses to obtain a higher activation score on the selected math-related neurons than dispreferred responses.

\paragraph{Reference model.}
The reference model is used only for computing the DPO log-probability margin. It does not participate in the neuron reward computation. The neuron reward is computed from the current policy model and remains differentiable with respect to the policy parameters.

\paragraph{Fixed neuron set.}
The selected math-related neuron set $\mathcal{N}_{\text{math}}$ is kept fixed throughout preference optimization. This makes different training runs comparable and avoids introducing extra instability from dynamically changing neuron sets.

\section{Training Dynamics}
\label{app:training_dynamics}

\begin{figure*}[t]
    \centering
    \includegraphics[width=0.98\textwidth]{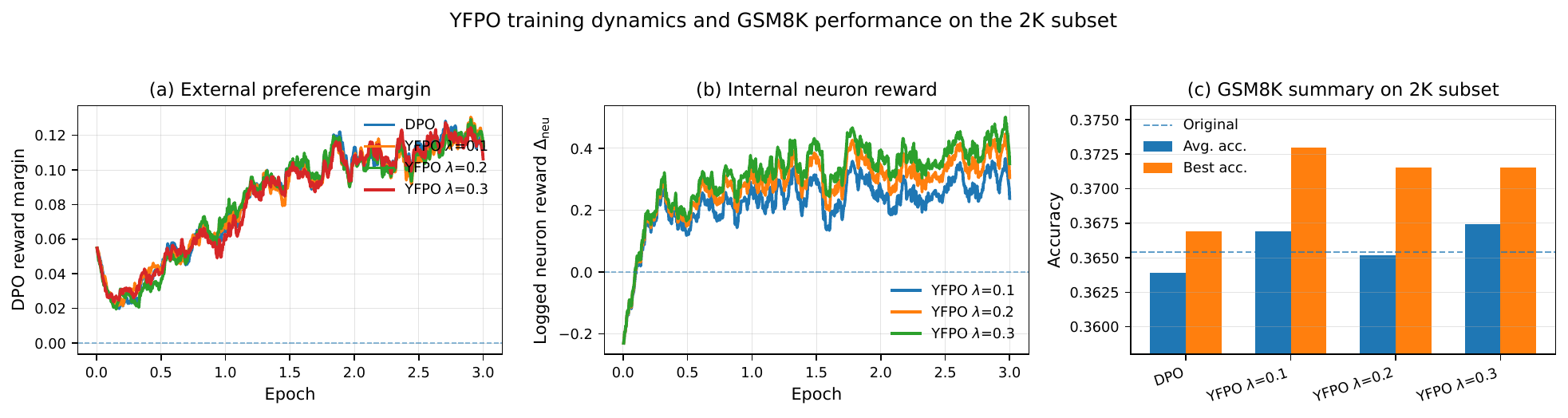}
    \caption{
    Training dynamics and GSM8K performance of YFPO on the 2K data subset.
    Left: external DPO preference margin during training.
    Middle: logged internal neuron reward for different $\lambda$ values.
    Right: average and best GSM8K accuracy across three checkpoints.
    YFPO preserves the external preference optimization signal while introducing a distinct neuron-level optimization signal, and achieves better best-checkpoint accuracy than DPO under multiple $\lambda$ settings.
    }
    \label{fig:yfpo_2k_dynamics}
\end{figure*}

Figure~\ref{fig:yfpo_2k_dynamics} shows the training dynamics of YFPO on the 2K subset. The external DPO preference margin follows broadly similar trends across DPO and YFPO variants, suggesting that the neuron reward does not collapse the original preference optimization process. Meanwhile, the internal neuron reward exhibits different trajectories across $\lambda$ values, indicating that it acts as an additional optimization signal rather than a static explanatory metric. These dynamics support the main results: YFPO can improve best-checkpoint accuracy under multiple reward strengths, but its effect remains sensitive to reward scale and training stage.

\section{Additional Training Details}
\label{app:training_details}

We report results from three saved training checkpoints, denoted as C1, C2, and C3 in the main text. The average accuracy reflects overall checkpoint-level behavior, while the best accuracy reflects the strongest observed checkpoint under each setting. This reporting protocol is used because YFPO shows visible checkpoint-level variance under different reward strengths.

\paragraph{Training configuration.}
Table~\ref{tab:training_config} summarizes the main training configuration. We train Qwen2-0.5B with bfloat16 precision and use the sigmoid DPO loss. The random seed and data seed are both set to 42. Checkpoints are saved at the end of each epoch, yielding C1, C2, and C3. For YFPO, we use $K=1024$ selected math-related neurons and sweep $\lambda$ over the values reported in the main tables. Unless explicitly specified, other DPO-related hyperparameters follow the default values of \texttt{DPOConfig}. We use the default values in the TRL version used in our experiments.

\begin{table}[t]
\centering
\small
\renewcommand{\arraystretch}{1.15}
\setlength{\tabcolsep}{5pt}
\begin{tabular}{ll}
\toprule
Hyperparameter & Value \\
\midrule
Base model & Qwen2-0.5B \\
Training precision & bfloat16 \\
Random seed & 42 \\
Data seed & 42 \\
Training epochs & 3 \\
Save strategy & per epoch \\
Batch size / device & 4 \\
Gradient accumulation & 2 \\
Effective batch size / device & 8 \\
Learning rate & $1\times10^{-6}$ \\
Max sequence length & 4096 \\
DPO loss type & sigmoid \\
DPO $\beta$ & 0.1 \\
Optimizer & AdamW (Transformers/TRL default) \\
Selected neurons $K$ & 1024 \\
YFPO coefficient $\lambda$ & reported in Tables~\ref{tab:gsm8k_10k} and~\ref{tab:gsm8k_2k} \\
Evaluation & GSM8K, greedy decoding \\
\bottomrule
\end{tabular}
\caption{
Training configuration used for DPO and DPO+YFPO experiments.
}
\label{tab:training_config}
\end{table}

The 2K subset used in Table~\ref{tab:gsm8k_2k} is a randomly sampled subset of \texttt{xinlai/Math-Step-DPO-10K}. It refers to a reduced training-data setting, not a random-neuron baseline. 

\end{document}